\documentclass[authoryear,3p, preprint,review,11pt]{elsarticle}

\usepackage[T1]{fontenc}
\usepackage{mathptmx}
\usepackage{verbatim}

\usepackage{enumitem}
\usepackage{multirow}
\usepackage{rotating}
\usepackage{epstopdf}
\usepackage{graphicx}
\usepackage{subfigure}
\usepackage{times}

\usepackage{makeidx}
\usepackage{multirow}
\usepackage{multicol}
\usepackage{graphicx}
\usepackage{epstopdf}
\usepackage{ulem}
\usepackage{epstopdf}

\usepackage{array}
\usepackage{amsmath,amssymb,amsfonts}
\usepackage{algorithmic}
\usepackage{algorithm}
\usepackage{graphicx}
\usepackage{textcomp}

\usepackage{epsfig}
\usepackage{makecell}
\usepackage{longtable,booktabs}
\usepackage{color}
\usepackage{longtable}
\usepackage{graphicx}
\usepackage{geometry}

\usepackage{longtable}
\usepackage{multirow}
\usepackage{array}
\usepackage{geometry} 
\usepackage{longtable} 
\usepackage{multirow} 
\usepackage{makecell} 
\usepackage{array} 
\usepackage{ragged2e} 
\usepackage{caption}

\usepackage{natbib}
\usepackage{caption}
\usepackage[colorlinks,
linkcolor=black,       
anchorcolor=black,  
citecolor=black,        
]{hyperref}

\journal{Preprint}

\begin{document}
	\begin{frontmatter}
		
		
		
		\title{Statistically Accurate and Robust Generative Prediction of Rock Discontinuities with A Tabular Foundation Model}
		
		\author[label1]{Han Meng}
		\author[label1]{Gang Mei\corref{cor1}}
		\ead{gang.mei@cugb.edu.cn}
		\author[label3]{Hong Tian}
        \author[label1]{Nengxiong Xu}
        \author[label1,label2]{Jianbing Peng}
        \cortext[cor1]{Corresponding author}
		
		\address[label1]{School of Engineering and Technology, China University of Geosciences (Beijing), 100083, Beijing, China}
		\address[label2]{School of Geological Engineering and Geomatics, Chang'an University, 710064, Xi'an, China}
		\address[label3]{School of Engineering, China University of Geosciences (Wuhan)}

	\begin{abstract}
		
		Rock discontinuities critically govern the mechanical behavior and stability of rock masses. Their internal distributions remain largely unobservable and are typically inferred from surface-exposed discontinuities using generative prediction approaches. However, surface-exposed observations are inherently sparse, and existing generative prediction approaches either fail to capture the underlying complex distribution patterns or lack robustness under data-sparse conditions. Here, we proposed a simple yet robust approach for statistically accurate generative prediction of rock discontinuities by utilizing a tabular foundation model. By leveraging the powerful sample learning capability of the foundation model specifically designed for small data, our approach can effectively capture the underlying complex distribution patterns within limited measured discontinuities. Comparative experiments on ten datasets with diverse scales and distribution patterns of discontinuities demonstrate superior accuracy and robustness over conventional statistical models and deep generative approaches. This work advances quantitative characterization of rock mass structures, supporting safer and more reliable data-driven geotechnical design.

	\end{abstract}
		

	\end{frontmatter}
	\newpage



\section{Introduction}
\label{sec1}

In nature, a wide range of human engineering activities are conducted in rock masses, including mineral resource extractions, transportation tunnel excavations, hydropower constructions, and urban infrastructure developments (\cite{RN42,RN44}). As a fundamental geological medium for engineering constructions, rock masses exhibit highly complex structural characteristics. The internal discontinuity systems (e.g., joints and fractures) are extensively developed and show considerable complexity and uncertainty in spatial distributions (\cite{RN41,RN43,RN55}). Among these discontinuities, some exhibit deterministic orientations governed by tectonic controls, while others display significant stochastic features as a consequence of long-term geological processes such as diagenesis, weathering, erosion, and redistribution of in-situ stress (\cite{RN48}). The spatial heterogeneity and variability of these discontinuities not only govern the overall mechanical behavior and stability of rock masses, but also influence all engineering activities in natural environments. Without an accurate characterization of the inherent distribution patterns, construction projects inevitably lead to unpredictable risks. Therefore, accurately characterizing the inherent distribution patterns of rock mass discontinuities is crucial for ensuring engineering safety and optimizing design, serving as a necessary prerequisite for addressing potential engineering risks (\cite{RN45}).

The internal distributions of rock discontinuities remain largely unobservable and are typically inferred from surface-exposed discontinuities using prediction approaches.  Discontinuities within rock masses can generally be categorized into two types: deterministic and stochastic discontinuities. Deterministic discontinuities are typically large-scale and clearly exposed, allowing us to directly measure them through field surveys (\cite{RN48}). In contrast, stochastic discontinuities are widely developed within rock masses under the long-term influence of tectonic activities and near-surface geological processes. These stochastic discontinuities are generally numerous and grouped, and their complex intersections and cuttings create a large number of rock blocks, imparting heterogeneity, discontinuity, and anisotropy to rock masses (\cite{RN12,RN46}). Such characteristics weaken rock mass integrity, create structural weak zones, and consequently establish stochastic discontinuities as the key factors controlling overall stability. Moreover, the inherent stochastic characteristics of these discontinuities reconfigure stress-transfer paths, induce stress concentration, and increase uncertainties in stability assessment. The abundance and concealed spatial distribution of stochastic discontinuities prevent their complete characterization by direct observation, thereby requiring indirect inference from limited surface-exposed observations (\cite{RN52,RN53,RN48}). This enhances the reliability of major infrastructure and effectively reduces engineering risks, while enabling potential geological hazards to be quantitatively assessed and systematically managed to ensure the safety and sustainable development of human activities. 

The prediction of stochastic discontinuities within rock masses is typically conducted using statistical approaches (\cite{RN52,RN64,RN65}). The underlying principle of these approaches is based on stochastic sampling and probabilistic statistics for simulating parameters such as discontinuity orientation, trace length, and density, thereby generating internal discontinuities whose statistical characteristics are consistent with those measured on rock surfaces.One of the most widely used statistical approaches is the Monte Carlo method (\cite{RN58,RN59,RN61,RN62}), which simplifies the distributions of geometric parameters into individual probability distribution functions (e.g., exponential, normal, or lognormal distributions). Probability distribution models are established from factually measured discontinuities to determine distribution types and corresponding statistical parameters (mean and variance). Subsequently, extensive repeated random sampling is performed according to the distribution function of each parameter until the generated parameters reproduce statistical features consistent with the measured discontinuities. Numerous studies have demonstrated that discontinuity parameters exhibit significant interdependence (\cite{RN66,RN67}). For example, aperture commonly shows a strong positive correlation with trace length, and dip direction is often coupled with dip angle. Monte Carlo method, however, generates each parameter independently, which neglects these inherent correlations and inevitably produces synthetic discontinuities whose statistical patterns deviate from those observed discontinuities.

The prediction of stochastic discontinuities within rock masses can also be achieved through generative deep learning approaches (\cite{RN68}). Various deep generative models—such as generative adversarial networks (GAN), denoising diffusion probabilistic models (DDPM), and variational autoencoders (VAE)---have been applied to predict discontinuities within rock masses. The core idea is to leverage the powerful nonlinear fitting capacity of deep neural networks to automatically learn the inherent patterns underlying observed discontinuities, and generate new data that are consistent with the statistical characteristics of observed discontinuities (\cite{RN69,RN70}). These deep generative model-based methods avoid simplifying the distributional characteristics of discontinuities and eliminate the need to prescribe them as fixed and explicit probabilistic forms. Instead, it can automatically capture the inherent complex distribution patterns of the discontinuities. However, deep generative models rely on sufficient and high-quality training datasets, while the limited surface---exposed discontinuities---owing to their inherent sparsity—fail to satisfy the data requirements for effective training of deep generative models. Such data scarcity forms a major constraint that significantly limits the performance of deep generative models.

Surface-exposed observations of rock discontinuities are inherently sparse, and existing generative prediction approaches either fail to capture the underlying complex distribution patterns or lack robustness under data-sparse conditions. Conventional statistical approaches simplify parameter distributions and assume independence among simulated parameters, which limits their capacity to represent the true complexity of discontinuities within rock masses. Generative deep learning approaches depend on sufficient and high-quality training samples and involve complex training parameter tuning. The limited number of surface-exposed discontinuities available falls far short of satisfying the training requirements for generative deep learning, severely limiting model performance.

Recently, foundation models such as TabPFN (\cite{RN23}) have specifically been designed for accurate prediction on small tabular data, which can effectively capture the underlying complex distribution patterns within limited observations. Owing to their large-scale pretraining, these models can be applied directly to generative tasks without the need for dataset-specific retraining or extensive hyperparameter tuning---constraints that commonly affect deep generative models (\cite{RN71,RN72,RN73}). Moreover, the broad prior knowledge learned from millions of tabular datasets endows foundation models with strong generalization capabilities, allowing them to generate diverse and statistically consistent discontinuity samples even under data-sparse conditions in rock masses.

Here, we proposed a simple yet robust approach for statistically accurate generative prediction (i.e., generation) of rock discontinuities by utilizing a tabular foundation model. The internal distributions of stochastic discontinuities within rock masses remain largely unobservable and are typically inferred from surface-exposed discontinuities using generative prediction approaches. However, surface-exposed observations are inherently sparse, and existing generative prediction approaches either fail to capture the underlying complex distribution patterns or lack robustness under data-sparse conditions. Foundation models specifically designed for accurate prediction on small data can effectively capture the underlying complex distribution patterns within limited observations. Inspired by the capability of the foundation model to capture complex distributional patterns from limited observations, we propose a simple yet robust approach for statistically accurate generative prediction (i.e., generation) of rock discontinuities. The approach employs TabPFN, a recently developed foundation model that is specifically designed for accurate prediction on small data. TabPFN integrates prior-posterior inference, pre-trained on extensive synthetic datasets to embed rich distributional priors. This architecture allows it to effectively transfer statistical knowledge to downstream tasks, significantly reducing reliance on sample size and ensuring accurate generation even under sparse-sample conditions. By leveraging the powerful sample learning capability of the TabPFN, specifically designed for small data, our approach can effectively capture the underlying complex distribution patterns within limited measured discontinuities.

\begin{figure}[H]
	\centering
	\includegraphics[width=\textwidth]{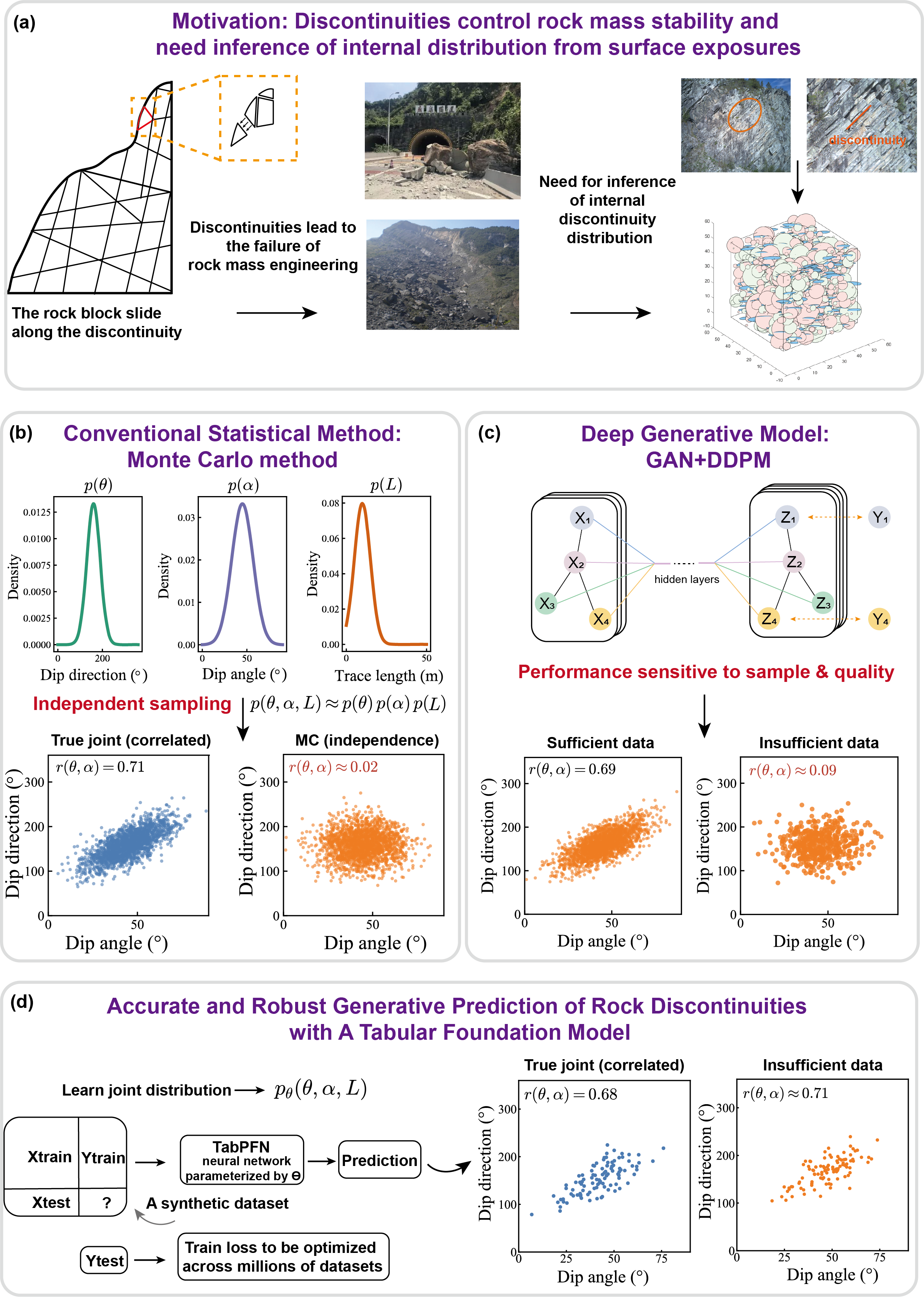}
	\caption{Accurate and robust generative prediction of rock discontinuities with a tabular foundation model}
	\label{fig:1}       
\end{figure}

To validate the accuracy and generalization capability of the proposed approach for generating stochastic discontinuities, we conducted comparative experiments on ten datasets with diverse scales and distribution patterns of discontinuities. These datasets encompass a wide range of discontinuity characteristics, ensuring the representativeness of the comparative results. The proposed approach was comprehensively compared with conventional statistical approaches (e.g., Monte Carlo method) and generative deep learning approaches (e.g., GANs and DDPMs). The evaluation metrics encompassed both univariate and multivariate aspects, including the statistical properties of individual discontinuity parameters, distributional similarity measures, and overall multivariate distribution similarity. In addition, parameter correlations were assessed through Pearson correlation coefficients.The schematic of the proposed method is shown in Fig.\ref{fig:1}

The comparative results demonstrate that the proposed approach achieves substantially higher consistency with measured discontinuities than conventional statistical approaches and deep generative model-based approaches. Moreover, the proposed approach demonstrates strong generalization and robustness, maintaining stable generation performance across datasets with varying patterns and quantities of discontinuities, significantly enhancing the reliability of stochastic discontinuity generation within rock masses. This work advances quantitative characterization of rock mass structures, supporting safer and more reliable data-driven geotechnical design. 

\section{Methods}
\label{sec:2}
\subsection{Approaches for generative prediction (generation) of rock discontinuities}
\label{sec:2.1}
\subsubsection{Conventional statistical approach: the Monte Carlo method}
\label{sec:2.1.1}

Statistical approaches generate discontinuity parameters such as orientation, trace length, spacing, and density through stochastic sampling and probabilistic modeling, thereby generating discontinuities whose statistical characteristics are consistent with those of the field-observed discontinuities. Fundamentally, these approaches rely on probability-based models to reproduce the distribution patterns of field-observed discontinuities.

Among statistical approaches, the Monte Carlo method is the most widely used (\cite{RN20,RN22}). It characterizes the key geometric parameters of discontinuities using independent probabilistic distribution functions, typically assuming forms such as negative exponential, normal, or log-normal distributions. Based on field-observed discontinuities, a probabilistic model is constructed for each parameter by defining its distribution type and associated statistical properties (e.g., mean and variance). Stochastic sampling is repeatedly performed according to these probabilistic models, and the statistical characteristics of the generated discontinuities are most consistent with the statistical characteristics of the field-observed discontinuities.

\subsubsection{Generative deep learning-based approach: GAN and DDPM}
\label{sec:2.1.2}

The prediction of stochastic discontinuities within rock masses can also be achieved through generative deep learning methods. Recent studies have employed deep generative models such as GANs and DDPMs to incorporate correlations among parameters of discontinuities, achieving significantly improved performance compared to the Monte Carlo method (\cite{RN18,RN19}). Other related work includes conditional GAN methods for digital rock reconstruction, enabling the generation of samples conditioned on specified distribution statistics (\cite{RN17}), methods for synthesizing three-dimensional geological facies models using deep generative networks (\cite{RN30}), and GAN-based approaches for geological parameterization and synthesis capable of reproducing highly complex geological patterns (\cite{RN31}).

Deep generative models (DGMs) leverage their exceptional capacity for nonlinear mapping to enable effective data-driven discontinuity generation. The essential mechanism involves transforming stochastic noise drawn from a simple initial distribution into complex, high-dimensional data structures. This is achieved through a hierarchical process of linear transformations followed by nonlinear activations within a multi-layer network. By repeatedly applying these linear-nonlinear mappings, DGMs effectively learn the underlying probability distribution of the training samples, enabling them to produce highly realistic synthetic realizations. When applied to discontinuity generation, DGMs learn the joint probabilistic structure embedded in field-observed discontinuities by training on the discontinuity samples. Once trained, the DGMs generate new discontinuities by drawing latent variables from the initial distribution and passing them through the learned network mapping. This enables DGMs to produce discontinuities that are statistically consistent with the field-observed discontinuities while retaining their inherent variability.

\vspace{1.2ex}
\noindent\hspace*{2em}\raisebox{0.2ex}{\textbullet}~\textbf{GAN}
\vspace{1.2ex}

The concept of GAN was first introduced by Goodfellow et al. at the NIPS conference in 2014 (\cite{RN40}). The theoretical foundation of GANs stems from the Nash equilibrium in game theory, where two competing agents reach a state in which neither can improve its outcome by unilaterally changing its strategy. The key innovation of GANs lies in integrating a generative model with a discriminative model that are jointly optimized through an adversarial learning process, substantially advancing the effectiveness of generative modeling.

The GAN comprises two fundamental components: a generator $G$ and a discriminator $D$. The generator produces synthetic samples, whereas the discriminator evaluates the authenticity of the input samples. During training, a random noise vector drawn from a latent space, which typically follows a predefined prior distribution such as a Gaussian, is fed into the generator. The generator then transforms this noise vector into a new sample $G(z)$ that closely resembles real data. Meanwhile, the discriminator receives both $G(z)$ and the real samples $X$, learning to distinguish between them. Functionally, the discriminator acts as a binary classifier that outputs the probability of a sample being real rather than generated. An output value approaching 1 indicates a higher likelihood of authenticity, whereas values closer to 0 suggest that the sample originates from the generator. Through this adversarial interaction, the generator progressively improves its ability to produce realistic data indistinguishable from real samples.

During training, the generator and discriminator engage in an adversarial optimization process. The discriminator seeks to distinguish real samples from those generated by the model, while the generator aims to produce synthetic data that are indistinguishable from real data. The adversarial relationship between the two networks is formulated by the loss function defined in  Eq. (\ref{eq1}). Training is performed alternately: the discriminator is updated first and the generator subsequently, forming an iterative process in which both networks are progressively refined. When optimizing the discriminator, the generator parameters are kept fixed, and the objective is to maximize the function described in Eq. (\ref{eq22}).

\begin{equation}\label{eq1}
	\min_{G}\max_{D} V(D,G) =
	\mathbb{E}_{\mathbf{x}\sim p_{\text{data}}(\mathbf{x})}[\log D(\mathbf{x})]
	+
	\mathbb{E}_{\mathbf{z}\sim p_{\mathbf{z}}(\mathbf{z})}[\log(1-D(G(\mathbf{z})))]
\end{equation}

\begin{equation}\label{eq22}
	\max_{D}\;
	\mathbb{E}_{\mathbf{x}\sim p_{\text{data}}(\mathbf{x})}[\log D(\mathbf{x})]
	+
	\mathbb{E}_{\mathbf{z}\sim p_{\mathbf{z}}(\mathbf{z})}[\log(1-D(G(\mathbf{z})))]
\end{equation}

During the generator training stage, the parameters of the discriminator are kept fixed. The generator aims to maximize the likelihood that the discriminator classifies generated samples as real, expressed as $D(G(z))$. In practice, this objective is achieved by minimizing the loss term $\log\bigl[1-D(G(z))\bigr]$. As training proceeds, the model gradually approaches a Nash equilibrium, where the discriminator can no longer effectively distinguish between synthetic and real samples. At this equilibrium, the predicted probabilities of the discriminator for both classes converge toward 0.5, indicating that the adversarial training process has stabilized and can be considered complete.

\vspace{1.2ex}
\noindent\hspace*{2em}\raisebox{0.2ex}{\textbullet}~\textbf{DDPM}
\vspace{1.2ex}

The DDPM represents a newly emerging class of deep generative models, with its conceptual foundation rooted in nonequilibrium thermodynamics (\cite{RN38}). The original framework was introduced by Sohl-Dickstein et al. in 2015 (\cite{RN39}), establishing the theoretical basis for diffusion modeling, and was later refined by Jonathan Ho et al. in 2020 through integration with deep learning techniques. Inspired by nonequilibrium thermodynamics, DDPM operates by progressively adding stochastic noise to data through a Markov chain—known as the forward diffusion process—and then training a neural network to learn the reverse process, or denoising pathway, to recover the underlying data distribution as the noise is gradually removed. The DDPM framework consists of three principal components: the forward diffusion process, the reverse denoising process, and the training procedure.

The first stage is the forward diffusion process, which performs a sequence of diffusion steps following a Markov chain. Over a predefined number of time steps, noise is incrementally added to the original sample~$X_0$, with each step corresponding to a different noise level. 
This progressive perturbation gradually transforms the original data distribution into a simple standard Gaussian distribution. The noise addition process can be expressed as a conditional normal distribution, defined by the following equation(Eq.(\ref{eq33})):

\begin{equation}\label{eq33}
	q(\mathbf{x}_t \mid \mathbf{x}_{t-1})
	= \mathcal{N}\bigl(\mathbf{x}_t;\; \sqrt{1-\beta_t}\,\mathbf{x}_{t-1},\; \beta_t\mathbf{I}\bigr)
\end{equation}

In this formulation, $q$ denotes the conditional probability distribution that characterizes the noise addition process from the previous state~$x_{t-1}$ to the current state~$x_t$ within each time step. The parameter~$\beta_t$ represents a predefined noise coefficient, and $\mathbf{I}$ denotes the identity matrix. Given the state~$x_{t-1}$ from the preceding time step, the sample~$x_t$ at the current step follows a normal distribution with both mean and variance defined by~$\beta_t$. The value of~$\beta_t$ increases with~$t$; as $t\to\infty$, the samples converge to an approximate Gaussian distribution.

The reverse process, which serves as the inverse of the forward diffusion procedure, can also be described as a Markov process. However, its transition probability distribution is unknown and must be approximated using a neural network. 
The initial state is sampled from a standard Gaussian distribution, $x_T\sim\mathcal{N}(\mathbf{0},\mathbf{I})$, and denoising is performed iteratively at each time step using a noise-prediction network~$\varepsilon_{\theta}(x_t,x_0)$. 
The process is governed by the following equation (Eq.(\ref{eq44})):

\begin{equation}\label{eq44}
	q(\mathbf{x}_{t-1} \mid \mathbf{x}_t, \mathbf{x}_0)
	= \mathcal{N}\bigl(\mathbf{x}_{t-1};\; \boldsymbol{\mu}_\theta(\mathbf{x}_t,\mathbf{x}_0),\; \beta_t\mathbf{I}\bigr)
\end{equation}

$\mu_{\theta}(x_t,x_0)$ represents the predicted mean of the sample distribution generated during the reverse diffusion process.

During training, the objective of the model is to minimize the difference between the predicted noise and the true noise. 
A sample~$x_0$ is drawn from the real data distribution, and Gaussian noise is added at a randomly selected timestep~$t$ to obtain~$x_t$. 
The network is then trained to predict the added noise~$\varepsilon$. 
Accordingly, the optimization objective can be simplified as follows (Eq.(\ref{eq55})):

\begin{equation}\label{eq55}
	\mathcal{L}(\theta)
	= \mathbb{E}_{\mathbf{x}_0,t,\boldsymbol{\varepsilon}}
	\Bigl[\bigl\|\boldsymbol{\varepsilon} - \boldsymbol{\varepsilon}_\theta(\mathbf{x}_t, t)\bigr\|^2\Bigr]
\end{equation}

The network parameters are optimized through gradient descent, progressively aligning the predicted noise with the true noise distribution. After training, the sampling process starts from Gaussian noise and performs iterative denoising steps, ultimately generating new samples that closely match the real data distribution.

\subsubsection{Foundation model-based approach: TabPFN}
\label{sec:2.1.3}

The foundation model specially designed for accurate prediction on small tabular data has recently emerged as a cutting-edge direction (~\cite{RN24}). These models are pretrained on large and diverse collections of datasets with the aim of learning general statistical representations rather than addressing a specific task. Such pretrained representations can then be adapted to a wide range of small-data generation or analysis tasks using only limited task-specific data or simple instruction-based adaptation. This pretrain-and-adapt paradigm reduces the reliance on task-specific hyperparameter tuning and alleviates long-standing challenges in traditional data generation, including significant task dependence and limited capability in sparse-sample scenarios (~\cite{RN25,RN26}).

TabPFN is one of the most advanced foundation models specifically developed for small data (~\cite{RN23}). It combines a Transformer architecture with the Prior-Data Fitted Networks Bayesian framework and is pretrained via meta-learning on approximately 120 million datasets. Through pretraining strategy, the foundation model acquires a form of algorithm-level generalization that enables effective prediction on small tasks without task-specific training. The strong representation learning and transfer capabilities acquired during pretraining also support effective end-to-end learning (~\cite{RN27}). Without manual feature engineering or task-specific preprocessing, the foundation model can directly learn mappings from raw inputs to outputs, reducing both the time and computational cost of developing generation models for new tasks while improving robustness. Compared with deep generative models such as GANs and DDPMs, TabPFN provides notable advantages, including stronger compatibility with tabular data structures, improved ability to capture feature dependencies, higher quality of generated samples, and more stable training behavior. In this study, TabPFN is employed to generate stochastic discontinuities by inferring new parameter values through a single forward pass, thereby transferring its pretrained distributional priors directly to the generation task. Through this mechanism, TabPFN generates discontinuities that remain statistically consistent with the distributional patterns of the field-observed discontinuities.

\subsection{Evaluation metrics}
\label{sec:2.2}
Comparative analyses were conducted to evaluate the performance of each approach after the discontinuities were generated using the three types of approaches. The analytical framework consists of two complementary components---qualitative assessment and quantitative evaluation---which together provide both clear visual comparison and scientifically rigorous, objective performance analysis.

\subsubsection{Qualitative assessment}
\label{sec:2.2.1}
The qualitative evaluation relies on visual comparison of the generated and field-observed discontinuities. Probability distribution histograms are used to compare the distributional shape, peak position, and tail behavior of each parameter. Boxplots are employed to examine differences in median, interquartile range, and outlier distribution, enabling an assessment of the dispersion of the generated discontinuities relative to the field-observed discontinuities. Two-dimensional scatter plots of dip direction and dip angle are further used to visualize their distribution and reveal the correlation structure between these parameters.

\subsubsection{Quantitative evaluation}
\label{sec:2.2.2}
For the quantitative evaluation, a set of statistical metrics was employed to assess the consistency between the generated discontinuities and the field-observed discontinuities. To evaluate individual parameters, the mean and standard deviation of dip direction, dip angle, and trace length were calculated to quantify differences in central tendency and dispersion. Distributional similarity was further assessed using the Kolmogorov-Smirnov test (KS test), the Wasserstein distance, and the chi-square goodness-of-fit test (chi-square test), providing complementary perspectives on discrepancies in cumulative distribution functions, distributional shape, and frequency counts. To analyze distribution patterns, correlation matrices, root-mean-square error (RMSE), mean absolute error (MAE), PERMANOVA statistics, and the Wasserstein distance were utilized to quantify differences. Moreover, Pearson correlation coefficients were calculated to examine whether each approach captures parameter correlations, particularly between dip direction and dip angle. These quantitative measures provide a comprehensive and objective evaluation of the performance of the different generative approaches.

\section{Results}
\label{sec:3}
We conducted comparative experiments on ten datasets encompassing diverse scales and distribution patterns of discontinuities. The experimental datasets were obtained from a publicly available datasets (~\cite{RN32}), which encompasses a wide range of discontinuity characteristics, ensuring the representativeness of the comparative results. The proposed approach was comprehensively compared with the conventional statistical approach (Monte Carlo method) and generative deep learning approaches (GAN and DDPM). The evaluation metrics encompassed both univariate and multivariate aspects, including the statistical properties of individual discontinuity parameters, distributional similarity measures, and overall multivariate distribution similarity. Moreover, parameter correlations were assessed through Pearson correlation coefficients.

The accurate characterization of discontinuity distributions within rock masses fundamentally depends on the reliable interpretation of the geometric parameter (~\cite{RN1981}). Discontinuity networks within rock masses are generally constructed using three probabilistic models: orientation distribution models, size distribution models, and spatial position distribution models (~\cite{RN22}). Therefore, in the comparative experiments, three key parameters of discontinuities, dip direction, dip angle, and trace length, are selected as the core parameters.

Surface-exposed observations of rock discontinuities are inherently sparse, and existing generative prediction approaches either fail to capture the underlying complex distribution patterns or lack robustness under data-sparse conditions. Accordingly, based on the sample size and distribution pattern of discontinuities, ten sets of discontinuities are categorized into the following four scenarios for detailed analysis. The distribution characteristics of each dataset are listed in Tab.\ref{tab:1}.

The comparative results demonstrate that, compared with conventional statistical approaches and deep generative model-based approaches, the proposed approach exhibits superior accuracy and robustness in stochastic discontinuity generation, exhibiting remarkable stability across all testing scenarios. It can effectively capture complex distributional structures and inherent parameter correlations. Under conditions of data-sparse and unclear statistical patterns, the proposed approach still generates discontinuities that are closely consistent with the observed discontinuities, highlighting its strong generalization capability and robust adaptability.

\begin{longtable}{%
		>{\centering\arraybackslash}p{3.6cm}
		>{\centering\arraybackslash}p{1.5cm}
		>{\centering\arraybackslash}p{9.0cm}}
     \caption{Summary of discontinuity datasets and their statistical characteristics}
     
	\label{tab:1} \\
	
	\toprule
	\textbf{Location} & \textbf{Number} & \textbf{Characteristics} \\
	\midrule
	\endfirsthead
	
	\toprule
	\textbf{Location} & \textbf{Number} & \textbf{Characteristics} \\
	\midrule
	\endhead
	
	\bottomrule
	\endfoot
	
	
	\multicolumn{3}{c}{\textbf{Discontinuities with relatively more samples and clear statistical regularity}} \\
	
	Oernlia, Norway & 766 &
	Dip direction and dip angle follow normal distributions, while trace length follows a log-normal distribution. A pronounced negative correlation exists between dip direction and dip angle. \\
	
	Laerdal, Norway & 562 &
	Dip direction and dip angle follow normal distributions, while trace length follows a log-normal distribution. A pronounced negative correlation exists between dip direction and dip angle. \\
	
	Thundovd, Norway & 325 &
	Dip direction and dip angle follow normal distributions, while trace length follows a log-normal distribution. A negative correlation exists between dip direction and dip angle. \\
	
	\midrule
	
	
	\multicolumn{3}{c}{\textbf{Discontinuities with relatively more samples and unclear statistical regularity}} \\
	
	Valle, Norway & 253 &
	Dip direction and dip angle follow normal distributions, while trace length follows a log-normal distribution. A negative correlation exists between dip direction and dip angle. \\
	
	\midrule
	
	
	\multicolumn{3}{c}{\textbf{Discontinuities with relatively less samples and clear statistical regularity}} \\
	
	Thundovd, Norway & 157 &
	Dip direction and dip angle follow normal distributions, while trace length follows a log-normal distribution. A positive correlation exists between dip direction and dip angle. \\
	
	Oernlia, Norway & 68 &
	Dip direction and dip angle follow normal distributions, while trace length follows a log-normal distribution. A positive correlation exists between dip direction and dip angle. \\
	
	\midrule
	
	
	\multicolumn{3}{c}{\textbf{Discontinuities with relatively less samples and unclear statistical regularity}} \\
	
	Valle, Norway & 40 &
	Dip direction and dip angle follow normal distributions, while trace length follows a log-normal distribution. A negative correlation exists between dip direction and dip angle. \\
	
	Chenjiazhuang, China & 104 &
	Dip direction and dip angle follow normal distributions, while trace length follows a log-normal distribution. A positive correlation exists between dip direction and dip angle. \\
	
	Chenjiazhuang, China & 115 &
	Dip direction and dip angle follow normal distributions, while trace length follows a log-normal distribution. No correlation exists between dip direction and dip angle. \\
	
	Chenjiazhuang, China & 119 &
	Dip direction and dip angle follow normal distributions, while trace length follows a log-normal distribution. No correlation exists between dip direction and dip angle. \\
	
\end{longtable}

\subsection{Experimental Scenario I: Discontinuities with relatively more samples and clear statistical regularity}
\label{sec:3.1}

For this experimental scenario, three datasets were selected for analysis: Oernlia slope (Group 1), Laerdal slope (Group 1), and Tundhovd slope (Group 2). These datasets are characterized by relatively more samples ($ \geq $300 samples per group) and exhibit clear statistical regularity.

\begin{figure}[H]
	\centering
	\includegraphics[width=\textwidth]{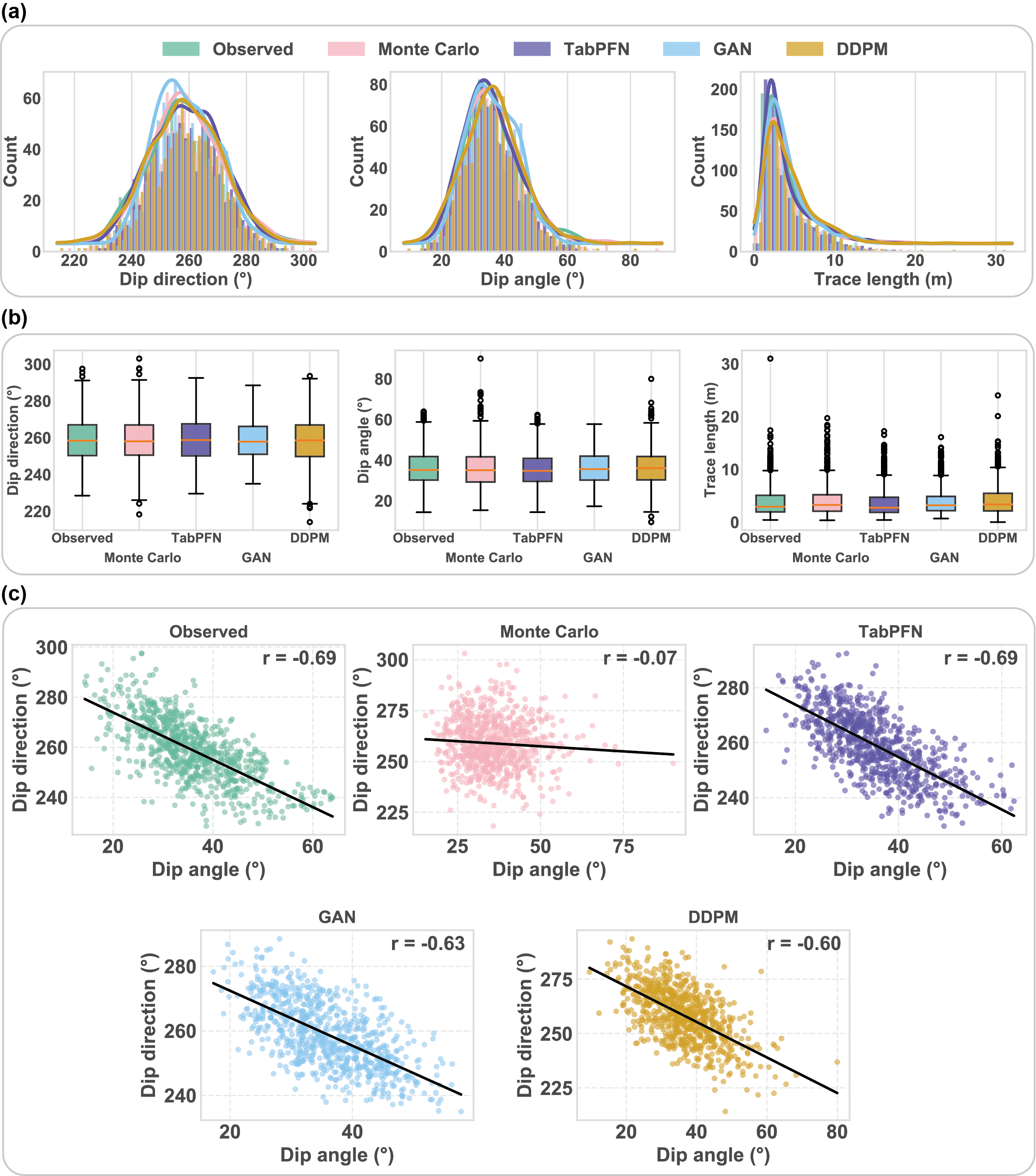}
	\caption{Experimental scenario I--example from the Oernlia Slope: comparison of observed and generated discontinuities using three types of approaches (histograms, box plots, and scatter plots).}
	\label{fig:2}       
\end{figure}

The first set of discontinuity from the Oernlia slope was selected as a representative case, and the results are illustrated in Fig.\ref{fig:2}. This dataset contains 677 discontinuities, in which dip direction and dip angle approximately follow normal distributions, while trace length follows a lognormal distribution. A pronounced negative correlation is observed between dip direction and dip angle, with a Pearson correlation coefficient of -0.69. The consistency between the generated and field-observed discontinuities was quantified and visualized using several comparative methods. These methods include probability histograms, box plots, and dip direction-dip angle scatter plots. Furthermore, the Pearson correlation coefficient was calculated to numerically quantify the statistical agreement between the synthetic and observed datasets. Overall, all approaches achieve strong consistency with the observed distributional patterns; however, the conventional statistical approach fails to capture the inherent correlations between dip direction and dip angle.

Based on the comparative results from the three discontinuity datasets, the conventional statistical approach performs well in reproducing distribution patterns of field-observations. For instance, it yields small relative errors in the mean and standard deviation of the trace length on the Laerdal slope and the dip angle on the Thundovd slope. However, its limitation is the inability to capture the inherent correlations among parameters, leading to the weakest performance on correlation-based metrics such as the Pearson correlation coefficient. More specifically, dip direction and dip angle of the generated discontinuities exhibit negligible correlation, indicating limited capability in reconstructing distribution patterns.

This limitation arises from the inherent principle of the approach. Its underlying mathematical principle resembles a naive Bayes approximation of a high-dimensional joint distribution, in which all parameters are assumed to be independent. During generation, the approach samples each marginal distribution separately and then combines the sampled values, thereby neglecting inherent statistical dependencies among parameters. When a parameter exhibits a simple distribution and weak correlations with others-as in the case of trace length on the Laerdal slope or dip angle on the Thundovd slope—the approach can reproduce its distribution patterns accurately. However, when implicit correlations exist among parameters, the independence assumption substantially weakens its ability to represent the distributional characteristics and to capture parameter correlations, resulting in inferior performance compared with the baseline approaches.

In contrast, deep generative model-based approaches exhibit strong capability in capturing complex distribution patterns, although their performance varies across datasets. The GAN-based approach demonstrates limited stability, with noticeable deviations from the observed discontinuities for certain parameters. The DDPM-based approach, by comparison, more effectively captures correlations and represents distribution characteristics. For example, in the Thundovd dataset, it achieves the lowest Frobenius norm, accurately reflecting the correlation between dip direction and dip angle. In the Laerdal dataset, it yields the smallest standard deviation errors for dip direction and dip angle among the four approaches. However, despite its advantages in correlation capturing, the DDPM-based approach remains less effective than the proposed foundation model-based approach in representing the distribution patterns of discontinuities.

The proposed foundation model-based approach achieves the best performance across all discontinuity datasets. It accurately represents the distribution characteristics of discontinuities, yielding smaller errors in mean and standard deviation and higher goodness-of-fit p-values than the baseline approaches. It also captures the inherent correlation between dip direction and dip angle with Pearson coefficients that closely match those of the observed discontinuities. Although its performance for certain parameters---such as trace length in the Laerdal dataset---is slightly inferior to that of the conventional statistical approach, its generative capability remains strong.

\subsection{Experimental Scenario II: Discontinuities with relatively more samples and unclear statistical regularity}
\label{sec:3.2}

The second set of discontinuities from the Valle slope was selected to evaluate the performance of different generative approaches under a scenario with relatively more samples, in which the distribution patterns of the parameters are unclear. This dataset contains a total of 253 discontinuities, with both dip direction and dip angle following approximately normal distributions, while trace length approximately follows a lognormal distribution. The Pearson correlation coefficient between dip direction and dip angle is -0.22, indicating a negative correlation between the two parameters.

\begin{figure}[H]
	\centering
	\includegraphics[width=\textwidth]{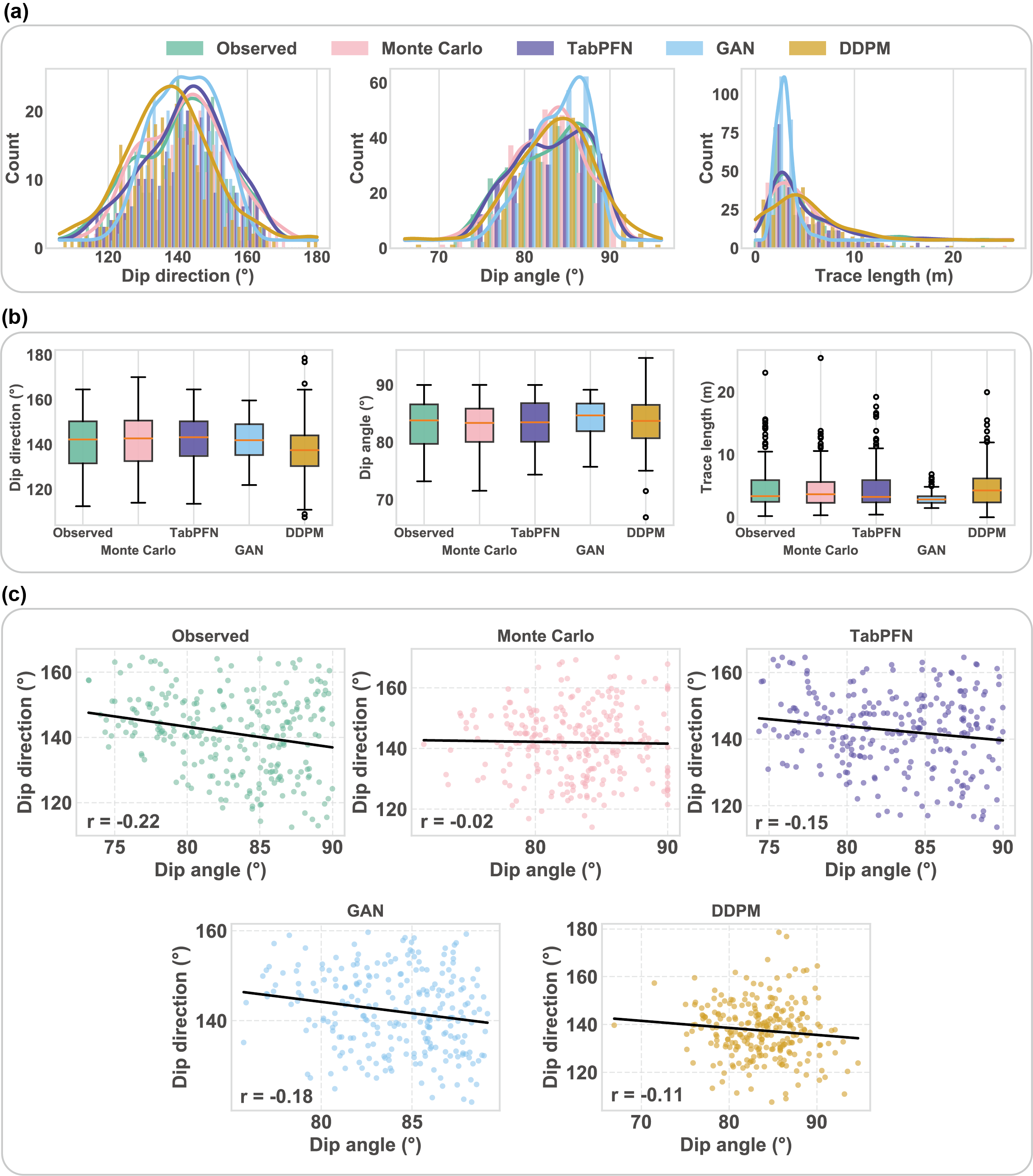}
	\caption{Experimental scenario II--example from the Valle Slope: comparison of observed and generated discontinuities using three types of approaches (histograms, box plots, and scatter plots).}
	\label{fig:3}       
\end{figure}

The histogram results illustrated in Fig.\ref{fig:3} demonstrate that the kernel density estimation (KDE) curves of discontinuities generated by both the conventional statistical approach and the proposed foundation model-based approach closely align with those of the field-observed discontinuities, whereas the deep generative model-based approaches exhibit significant deviations. Similarly, the boxplots indicate that the distributions of discontinuities generated by the conventional and proposed approaches correspond more closely to the observed discontinuities, while those generated by the deep generative model-based approach are noticeably more dispersed. Furthermore, the scatter plots reveal that the conventional statistical approach yields a Pearson correlation coefficient between dip direction and dip angle near zero, while the proposed foundation model-based approach and the GAN-based approach produce coefficients that more accurately reflect the correlation in parameters of the field-observed discontinuities.

In general, the conventional statistical approach exhibits limited performance. For dip direction, it yields relatively small errors in mean and standard deviation, and the fitted distributions pass both the KS and chi-square tests. However, the distribution patterns of the generated dip angle and trace length show noticeable deviations from the distribution patterns of the field-observed discontinuities, indicating weaker generation accuracy for these parameters. In terms of parameter correlation capturing, the conventional statistical approach performs the worst, producing a Frobenius norm of 0.331---the highest among all approaches---which reflects its limited ability to capture parameter correlations. These limitations become more pronounced when the underlying distribution patterns are not well defined. As the approach assumes parameter independence and characterizes each parameter with simplified probabilistic forms, it cannot represent discontinuities governed by complex distributions.

Although deep generative model-based approaches are capable of capturing correlations among discontinuity parameters, their accuracy in reproducing distributions patterns is noticeably weaker than that of both the conventional statistical method and the proposed foundation model-based approach. Although this dataset contains more discontinuities than the other groups, it remains a sparse-sample dataset for deep learning models. Under discontinuity conditions with relatively more samples and unclear statistical regularity, deep generative model-based approaches cannot effectively leverage their advantages and are prone to issues such as underfitting and mode collapse due to insufficient samples. As a result, they struggle to reproduce the distribution patterns of discontinuities and to reliably represent the underlying parameter correlations.

The proposed foundation model-based approach achieves the best performance on this dataset. It accurately captures the distribution patterns of discontinuities, producing the highest KS and chi-square p-values and the smallest Wasserstein distances among all approaches. These results indicate that the distributions of the generated discontinuities are most consistent with those of the field-observed discontinuities. In evaluating parameter correlations, the approach attains a Frobenius norm of only 0.171, substantially lower than those of both the conventional statistical approach and the deep generative model-based approaches. It also yields the highest PERMANOVA p-value, reflecting its ability to accurately capture complex inherent parameter correlations and provide a reliable representation of the statistical characteristics of the field-observed discontinuities.

\subsection{Experimental Scenario III: Discontinuities with relatively less samples and clear statistical regularity}
\label{sec:3.3}

For this experimental scenario, two datasets were selected for analysis: Oernlia slope (Group 3) and Thundovd slope (Group 1). These datasets are characterized by relatively less samples and exhibit clear statistical regularity.

\begin{figure}[H]
	\centering
	\includegraphics[width=\textwidth]{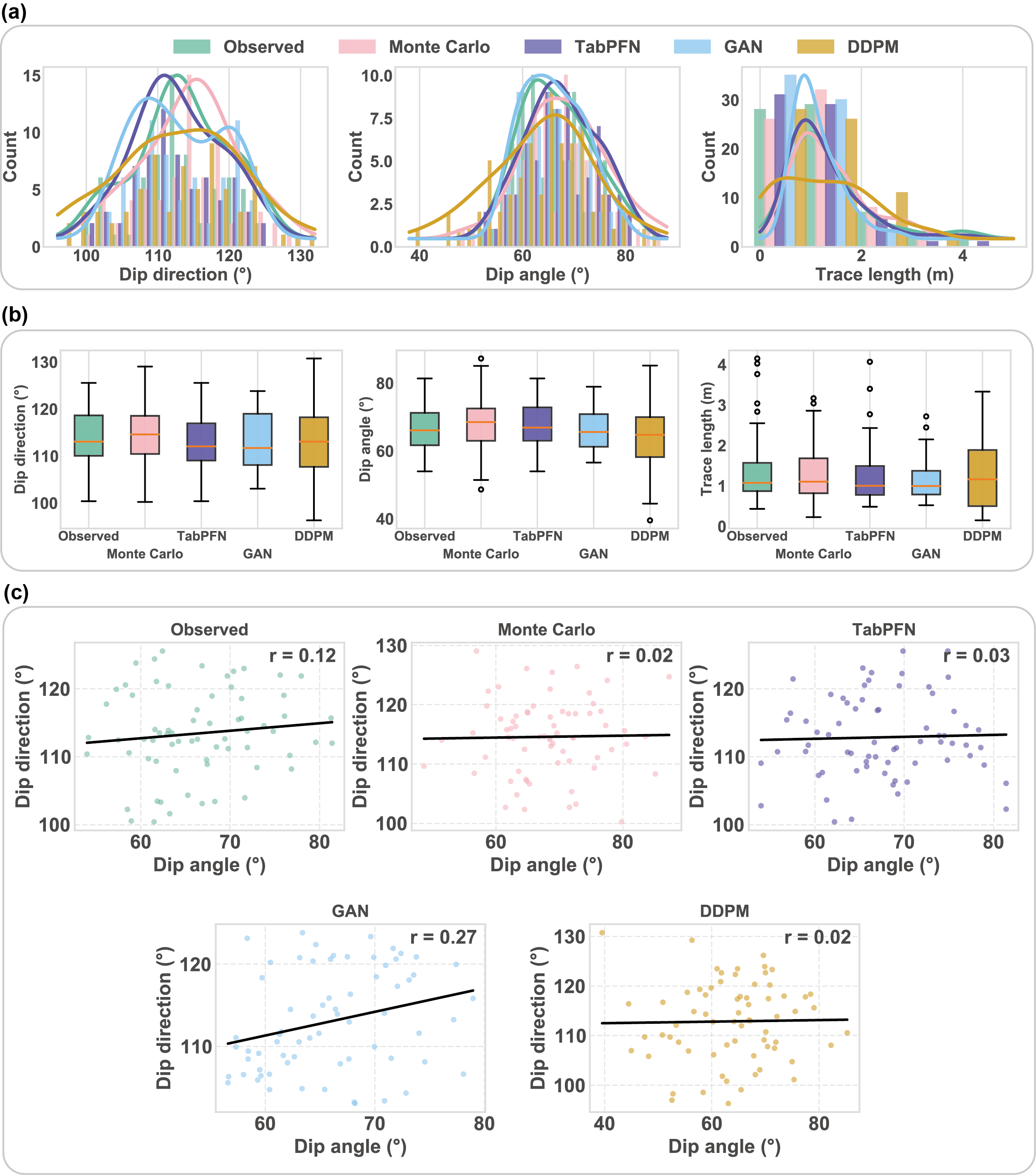}
	\caption{Experimental scenario III--example from the Oernlia Slope: comparison of observed and generated discontinuities using three types of approaches (histograms, box plots, and scatter plots)}
	\label{fig:4}       
\end{figure}

The third set of discontinuity from the Oernlia slope was selected as a representative case, and the results are illustrated in Fig. \ref{fig:4}. This dataset contains 68 discontinuities, in which dip direction and dip angle approximately follow normal distributions, while trace length follows a lognormal distribution. A weak positive correlation is observed between dip direction and dip angle, with a Pearson correlation coefficient of 0.12. The consistency between the generated and field-observed discontinuities was quantified and visualized using several comparative methods. These methods include probability histograms, box plots, and dip direction-dip angle scatter plots. Furthermore, the Pearson correlation coefficient was calculated to numerically quantify the statistical agreement between the synthetic and observed datasets. 

The histogram results illustrated in Fig. \ref{fig:4} demonstrate that the KDE curves of discontinuities generated by both the conventional statistical approach and the proposed foundation model-based approach closely align with those of the field-observed discontinuities, whereas the deep generative model-based approaches exhibit significant deviations. Similarly, the boxplots indicate that the distribution patterns of discontinuities generated by both the conventional and proposed approaches correspond more closely to the field-observed discontinuities, while those generated by the deep generative model-based approach are noticeably more dispersed. Furthermore, the scatter plots reveal that the conventional statistical approach, the proposed approach, and the DDPM-based approach are close to zero, while the correlation learned by the GAN-based approach deviates to some extent from the observed relationship between dip direction and dip angle in the discontinuities.

The comparative results from the two discontinuity datasets indicated that, the conventional statistical approach can accurately capture the distribution of parameters. However, when a strong correlation exists between dip direction and dip angle, its inability to capture inherent correlations leads to poor performance in representing the distribution patterns of discontinuities. As a result, it consistently yields relatively large Frobenius norm values, reflecting its weaker capability in distribution patterns reconstruction.

Deep generative model-based approaches perform better than the conventional statistical approach but remain inferior to the proposed foundation model-based approach. The GAN-based approach performs best on the Thundovd slope dataset, yielding the highest PERMANOVA p-value, which indicates the closest consistency between the distributions of the generated discontinuities and those of the field-observed discontinuities. On the Oernlia slope dataset, the GAN-based approach also performs well in fitting the distributions of dip direction and dip angle, although its accuracy for trace length is slightly lower than that of the proposed foundation model-based approach. In contrast, the DDPM-based approach demonstrates relatively limited performance. While it performs relatively well on certain metrics—such as achieving a lower Frobenius norm on the Thundovd dataset---it provides limited accuracy in reproducing the distribution patterns. With limited samples, the DDPM-based model fails to adequately capture the distribution characteristics.

The proposed foundation model-based approach demonstrates strong robustness and generalization capability, exhibiting robust performance on two discontinuity datasets. On the Thundovd slope dataset, it provides the most accurate generation for trace length and maintains stable performance across all evaluation metrics. On the Oernlia slope dataset, it provides the most accurate generation for all parameters, with PERMANOVA p-values higher than those of the other methods. These results indicate that the proposed foundation model-based approach is able to capture the distribution characteristics of the discontinuities and remains robust under sparse-sample datasets.

\subsection{Experimental Scenario IV: Discontinuities with relatively less samples and unclear statistical regularity}
\label{sec:3.4}

\begin{figure}[H]
	\centering
	\includegraphics[width=\textwidth]{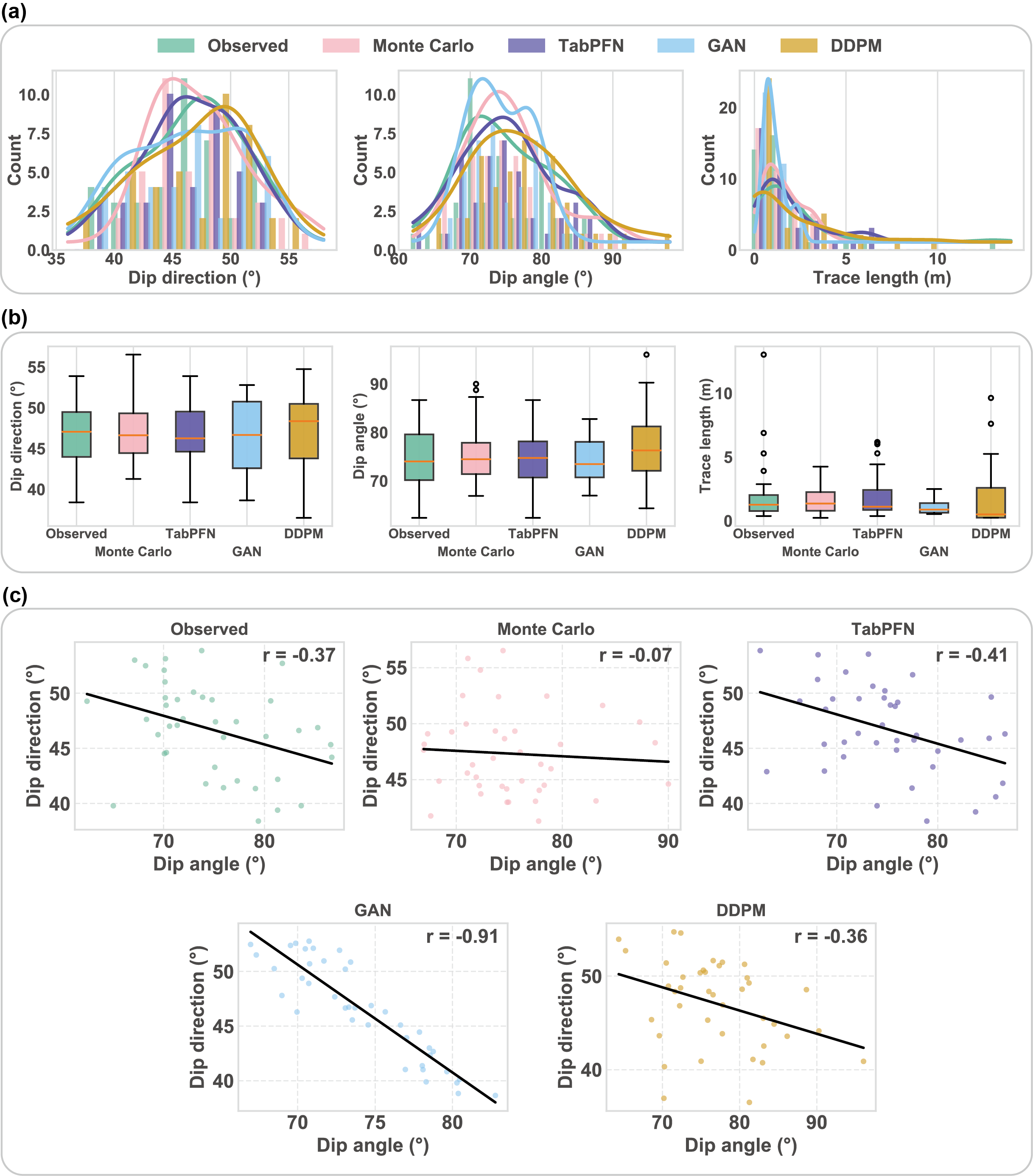}
	\caption{Experimental scenario IV--example from the Valle Slope: comparison of observed and generated discontinuities using three types of approaches (histograms, box plots, and scatter plots)}
	\label{fig:5}       
\end{figure}

For this experimental scenario, four datasets were selected for analysis: Valle slope (Group 3) and three datasets of Chenjiazhuang slope. These datasets are characterized by relatively less samples and exhibit unclear statistical regularity. For clarity of presentation, the third dataset from the Valle slope is selected as a representative example, and the corresponding results are shown in Fig. \ref{fig:5}. This dataset contains 40 discontinuities. Both dip direction and dip angle approximately follow normal distributions, whereas trace length follows a lognormal distribution. The Pearson correlation coefficient between dip direction and dip angle is -0.37, indicating a clear negative correlation between the two parameters.

The results illustrated in Fig. \ref{fig:5} demonstrate that the three types approaches exhibit clear differences in their generation performance. The proposed foundation model-based approach provides the most accurate generation, with the KDE curves of the generated discontinuities closely consistent with the field-observed discontinuities, whereas the deviations produced by the other methods are more pronounced. The boxplots further indicate that the distribution patterns of the discontinuities generated by the conventional statistical approach and the proposed approach yield most similar to those of the field-observed discontinuities, while the distribution patterns generated by the deep generative model-based approaches are more widely spread. In terms of capturing parameter correlations, the proposed approach and the DDPM-based deep generative approach perform particularly well, producing Pearson correlation coefficients for dip direction and dip angle (0.41 and 0.36, respectively) that are closest to those of the field-observed discontinuities. By contrast, the GAN-based deep generative approach exhibits excessively strong correlations due to underfitting, resulting in deviations from the true parameter correlations.

The comparative results from both discontinuity datasets indicate that, the conventional statistical approach demonstrates acceptable performance in generation dip direction and dip angle but performs poorly for trace length, with clearly higher errors. The Frobenius norm is consistently yields the highest error across all experimental datasets, indicating the weakest performance in representing distribution patterns. This weakness lies in its inability to capture inherent correlations among parameters.

The deep generative model-based approaches yield lower accuracy than the conventional statistical approach, with the DDPM-based deep generative approach generally achieving better results than the GAN-based deep generative approach. The GAN-based approach exhibits the weakest performance across all experimental datasets. It produces noticeably higher fitting errors---for example, in the Valle dataset, its relative error for trace length is higher than those of the other approaches. In addition, it consistently yields low KS and chi-square p-values, while its Frobenius norms and Wasserstein distances are typically the highest, indicating limited capability in representing the distribution patterns.

Among the three types approaches, the proposed foundation model-based approach delivers the highest reliable and accurate performance on this dataset. For individual parameter, the generated values exhibit consistently low relative errors in both mean and standard deviation, indicating that the approach effectively captures the distribution characteristics of each parameter. In terms of distribution patterns, the proposed approach yields the lowest Frobenius norms for the first and third groups of the Chenjiazhuang dataset and similarly low values for the Valle dataset and the second Chenjiazhuang group. This demonstrates its strong capability in capturing the inherent correlations among parameters. Its PERMANOVA p-values are also generally the highest, indicating that the distribution patterns of the generated discontinuities most consistent with those of the field-observed discontinuities. Notably, even for the Valle dataset---containing only 40 discontinuities---the approach maintains high accuracy, highlighting its strong generalization ability and adaptability under sparse-sample conditions.

\section{Discussion}
\label{sec:4}

Rock discontinuities critically govern the mechanical behavior and stability of rock masses. Their internal distributions remain largely unobservable and are typically inferred from surface-exposed discontinuities using generative prediction approaches. However, surface-exposed observations are inherently sparse, as observable outcrops are limited and spatially restricted, which hinders the acquisition of sufficient discontinuity samples. Consequently, accurately characterizing the distribution patterns of discontinuities within rock masses under data-sparse conditions remains a critical challenge for ensuring engineering safety.

Here, we proposed a simple yet robust approach for statistically accurate generative prediction (generation) of rock discontinuities by utilizing a tabular foundation model. To validate the accuracy and generalization capability of the proposed approach, we conducted experimental analyses on ten datasets with diverse scales and distribution patterns of field-observed discontinuities. The results of the comparative analyses indicate that:

(1) The proposed approach demonstrates superior performance across all four experimental scenarios. It exhibits outstanding capability in reproducing single-parameter statistical characteristics, capturing multi-parameter correlations, and maintaining strong generalization and robustness when applied to sparse datasets.

(2) The conventional statistical approach performs satisfactorily in single-parameter distribution fitting; however, due to its inherent assumption of parameter independence, it neglects inherent correlations among parameters, resulting in limited applicability.

(3) The performance of deep generative model-based approaches is highly dependent on sample size and distribution regularity. While they achieve acceptable results when the observations are sufficiently numerous and exhibit clear distributional patterns, their generation stability and ability to reconstruct complex distributions deteriorate significantly under sparse-sample or unclear data conditions.

\subsection{Influence of sample size on approach performance}
\label{sec:4.1}

Comparative results demonstrate that, although the generative performance of all approaches improves as sample size increases, the extent of such improvement varies substantially across methods.

The sample size of field-observed discontinuities exerts only a modest influence on the performance of conventional statistical methods. Increasing the number of observations leads to slight improvements in the accuracy of single-parameter fitting, yet the gains remain limited compared with the results obtained under sparse-sample conditions. More importantly, conventional statistical methods are inherently unable to capture the correlations between discontinuity parameters. The Pearson correlation coefficient remains consistently close to zero across all sample sizes, indicating that this approach fails to recover the true joint distribution of discontinuity orientations and, consequently, cannot reproduce the underlying distributional patterns of the observed discontinuities.

This limitation arises from the mathematical nature of conventional statistical methods, which approximate the high-dimensional joint distribution of discontinuity parameters using a naive independence assumption. In practice, data are generated by independently sampling each parameter from its predefined marginal distribution. Once the form of these marginal distributions and their key statistics (e.g., mean and variance) are specified, increasing the sample size contributes little additional benefit. Moreover, by assuming statistical independence among parameters, the approach performs stochastic sampling separately for each marginal distribution, entirely ignoring inherent correlations among parameters. When such inherent correlations are present, the assumption of independence fails, leading the approach to reproduce only marginal behaviors rather than capturing the intrinsic structure of the discontinuity distribution as achieved by other approaches.

The sample size of field-observed discontinuities has the strongest impact on deep generative model-based approaches, which fundamentally depend on sufficient samples to learn complex, high-dimensional latent distributions. When sufficient samples are available, these models can be adequately trained to capture intricate nonlinear relationships and generate rock discontinuities that are statistically consistent with the observed discontinuities. In contrast, under sparse-sample conditions, their ability to learn meaningful representations becomes severely limited, making them prone to underfitting, unstable training dynamics, mode collapse, and poor generalization. As a result, the discontinuities generated under low-sample regimes deviate markedly from the distributional characteristics of the observed discontinuities.

For GANs, the data distribution is learned through an adversarial process between the generator and the discriminator. The generator attempts to produce synthetic samples indistinguishable from field-observations, while the discriminator seeks to distinguish real data from generated data. Achieving equilibrium in this adversarial process requires abundant and diverse training samples; when the sample dataset is small, the discriminator rapidly underfits to field-observed, preventing effective learning by the generator and ultimately causing mode collapse. 

DDPMs, in contrast, learn data distributions through a forward diffusion and reverse denoising process. Gaussian noise is progressively added to real samples to simulate diffusion, and a neural network is trained to iteratively remove noise and reconstruct data from pure noise. This denoising process demands large datasets to learn the complex mapping between noise and structure. With insufficient samples, DDPM fails to capture fine-grained or localized features, leading to inaccurate denoising and poor-quality synthetic discontinuities that cannot faithfully represent the true data distribution.

Both GANs-based and DDPMs-based deep generative approaches fundamentally depend on sufficient samples to learn the intrinsic statistical structure of discontinuities, and data scarcity severely limits their ability to capture the essential distribution characteristics of discontinuities.

The sample size of field-observed discontinuities exerts only a limited impact on the proposed foundation model-based approach. The foundation model, specifically designed for small tabular data, adopts an in-context learning framework in which both training and target samples are jointly fed into the model, allowing it to perform fitting and prediction directly during inference without additional gradient updates. This mechanism enables efficient adaptation to new sparse-sample tasks without retraining. Furthermore, the prior regularization inherent to the large tabular model naturally mitigates underfitting in small-sample scenarios. Consequently, when applied to discontinuity datasets, the proposed approach can efficiently and stably generate realistic data even with extremely limited samples (e.g., only 40 discontinuities in the Valle slope dataset).

\subsection{Influence of distribution regularity on approach performance}
\label{sec:4.2}

The distribution patterns of discontinuities also affect the performance of different generation approaches. Comparative results indicate that when the parameter distributions are relatively clear, all generation approaches produce discontinuities that closely consistent with the field-observed discontinuities. However, when the parameter distributions become unclear and are strongly affected by noise, the performance differences among the various approaches become substantially more pronounced.

The distribution patterns of discontinuities have the strongest impact on the conventional statistical approach. This approach represents explicit, data-driven generations that assume each discontinuity parameter follows a single probabilistic distribution. This approach generates discontinuities based on the mean and variance of each parameter and independently samples from assumed distributions until the generated discontinuities approximate the field-observed discontinuities. The performance of this approach is therefore highly sensitive to how accurately the underlying distribution can be represented.

When the distributions of discontinuity parameters are simple, follow clear single-probability forms, and exhibit weak parameter correlations, the conventional statistical approach performs well for distribution patterns. However, as distribution complexity increases---particularly under strong noise or when inherent correlations exist among parameters---the predefined statistical distribution assumptions cannot accurately represent the underlying characteristics. The dependence of the sampling process on the assumed probabilistic form prevents the approach from capturing inherent correlations. As a result, the distribution patterns of the generated discontinuities deviate from distribution patterns of the field-observed discontinuities. Under conditions of unclear distribution regularities, the performance gap between the conventional statistical approach and the deep generative model-based approaches or the proposed foundation model-based approach becomes increasingly pronounced.

The distribution patterns of discontinuities exert a noticeable impact on the deep generative model-based approaches. These approaches employ deep neural networks to implicitly approximate high-dimensional probability distributions. However, when the discontinuity distributions are highly complex or noisy, the training process becomes unstable. For instance, GAN-based models rely on stable gradient signals; under complex or sparse-sample conditions, the generator fails to obtain stable gradient feedback, often leading to mode collapse or gradient vanishing. As a result, the deep generative model cannot reproduce the distribution patterns, causing the distribution patterns of the generated discontinuities deviate from distribution patterns of the field-observed discontinuities. 
 
The distribution pattern of discontinuities has a much lower impact on the proposed foundation model-based approach. The foundation model is pretrained on large-scale tabular datasets that cover a wide range of distribution patterns, allowing it to acquire rich prior knowledge of statistical characteristic. When deal with data exhibiting complex distribution patterns, the foundation model can effectively transfer the learned inductive capabilities. Under the in-context learning framework, the limited observations and the target samples are jointly provided during inference, enabling the foundation model to perform prediction in a single forward pass without additional training. This process reduces the risk of underfitting under high-noise conditions and avoids training instabilities such as gradient instability or mode collapse. Consequently, the proposed foundation model-based approach maintains high accuracy and stability in generating discontinuities, even when applied to samples exhibiting complex stochastic patterns.

\subsection{Advantages and limitations of the proposed approach}
\label{sec:4.3}
Compared to other approaches, the foundation model-based approach proposed in this study demonstrates reliable consistent with the field-observed discontinuities. The foundation model combines pretrained distributional priors with an in-context learning mechanism. Large-scale pretraining on millions of tabular tasks that cover a wide range of distributional patterns allows foundation model to encode rich statistical priors. Leveraging these priors, the foundation model can infer statistically consistent distributional patterns even under sparse-sample conditions. During inference, TabPFN uses in-context conditional prediction to approximate the posterior distribution of discontinuity parameters in a single forward pass, without any task-specific training. These mechanisms explain why the proposed approach achieves higher accuracy than the baseline approaches.

Furthermore, the proposed approach exhibits strong generalization and robustness. It maintains stable performance across a wide range of distribution patterns and sample sizes, demonstrating significantly lower dependence on sample size and distribution patterns compared to baseline approaches. By performing conditional inference directly on the limited field-observations, the foundation model effectively transfers the statistical priors learned during pretraining to new tasks. This mechanism reduces the reliance on sample sizes and enables accurate generation under sparse-sample conditions, while avoiding the mode-collapse behaviors commonly observed in deep generative models when dealing with complex distribution patterns.

Meanwhile, the proposed approach offers a simple implementation. Unlike deep generative model-based approaches that rely on iterative training and extensive hyperparameter tuning, the proposed approach generates discontinuities in a single forward pass. This design significantly simplifies the modeling workflow and enhances reproducibility.

A primary limitation of the proposed foundation model-based approach is its lack of physical interpretability. Although the approach can accurately capture complex distributional patterns but lack geological priors. As a result, the generated discontinuities achieve statistical consistency but lack a clear physical basis.

\section{Conclusion}
\label{sec:5}

Accurately characterizing the intrinsic distribution patterns of discontinuities within rock masses is essential for ensuring construction safety, optimizing design, and mitigating engineering risks. However, the internal distributions of discontinuities within rock masses remain largely unobservable and are typically inferred from surface-exposed discontinuities using generative prediction approaches. Surface-exposed observations are inherently sparse, and existing generative prediction approaches either fail to capture the underlying complex distribution patterns or lack robustness under data-sparse conditions. 

To address this problem, we proposed a simple yet robust approach for statistically accurate generative prediction of rock discontinuities by utilizing a tabular foundation model.  Inspired by the capability of the foundation model to capture complex distributional patterns from limited observations, we propose a simple yet robust approach for statistically accurate generative prediction (generation) of rock discontinuities. More specifically, the foundation model TabPFN is employed to effectively capture the underlying complex distribution patterns within limited measured discontinuities by leveraging its powerful sample-learning capability specifically designed for small-data.

We conducted comparative experiments on ten datasets encompassing diverse scales and distributional patterns of discontinuities. The comparative results show that the proposed approach consistently achieves the best performance in stochastic discontinuity generation and maintains remarkable stability across all testing scenarios, whereas the other two approaches exhibit limitations. The main findings are summarized as follows:

(1) The proposed approach demonstrates superior performance across all four experimental scenarios. It exhibits outstanding capability in reproducing single-parameter statistical characteristics, capturing multi-parameter correlations, and maintaining strong generalization and robustness when applied to sparse datasets.

(2) The conventional statistical approach performs satisfactorily in single-parameter distribution fitting; however, due to its inherent assumption of parameter independence, it neglects inherent correlations among parameters, resulting in limited applicability.

(3) The performance of deep generative model-based approaches is highly dependent on sample size and distribution regularity. While they achieve acceptable results when the observations are sufficiently numerous and exhibit clear distributional patterns, their generation stability and ability to reconstruct complex distributions deteriorate significantly under sparse-sample or unclear data conditions.

This work advances quantitative characterization of rock mass structures, supporting safer and more reliable data-driven geotechnical design.

\section*{Declaration of competing interest}

The authors declare that they have no known competing financial interests or personal relationships that could have appeared to influence the work reported in this paper.

\section*{Acknowledgments}

The data source for this article is the structural plane data provided by Larissa Elisabeth Darwell, and we are very grateful. This work was supported by the National Natural Science Foundation of China (Grant No.42277161).

\newpage




\begin{thebibliography}{41}
	\expandafter\ifx\csname natexlab\endcsname\relax\def\natexlab#1{#1}\fi
	\providecommand{\url}[1]{\texttt{#1}}
	\providecommand{\href}[2]{#2}
	\providecommand{\path}[1]{#1}
	\providecommand{\DOIprefix}{doi:}
	\providecommand{\ArXivprefix}{arXiv:}
	\providecommand{\URLprefix}{URL: }
	\providecommand{\Pubmedprefix}{pmid:}
	\providecommand{\doi}[1]{\href{http://dx.doi.org/#1}{\path{#1}}}
	\providecommand{\Pubmed}[1]{\href{pmid:#1}{\path{#1}}}
	\providecommand{\bibinfo}[2]{#2}
	\ifx\xfnm\relax \def\xfnm[#1]{\unskip,\space#1}\fi
	\bibitem[{Borisov et~al.(2022a)Borisov, Leemann, Seßler, Haug, Pawelczyk and
		Kasneci}]{RN72}
	\bibinfo{author}{Borisov, V.}, \bibinfo{author}{Leemann, T.},
	\bibinfo{author}{Seßler, K.}, \bibinfo{author}{Haug, J.},
	\bibinfo{author}{Pawelczyk, M.}, \bibinfo{author}{Kasneci, G.},
	\bibinfo{year}{2022}a.
	\newblock \bibinfo{title}{Deep neural networks and tabular data: A survey}.
	\newblock \bibinfo{journal}{IEEE transactions on neural networks learning
		systems} \bibinfo{volume}{35}, \bibinfo{pages}{7499--7519}.
	\bibitem[{Borisov et~al.(2022b)Borisov, Leemann, Seßler, Haug, Pawelczyk and
		Kasneci}]{RN24}
	\bibinfo{author}{Borisov, V.}, \bibinfo{author}{Leemann, T.},
	\bibinfo{author}{Seßler, K.}, \bibinfo{author}{Haug, J.},
	\bibinfo{author}{Pawelczyk, M.}, \bibinfo{author}{Kasneci, G.},
	\bibinfo{year}{2022}b.
	\newblock \bibinfo{title}{Deep neural networks and tabular data: A survey}.
	\newblock \bibinfo{journal}{IEEE transactions on neural networks learning
		systems} \bibinfo{volume}{35}, \bibinfo{pages}{7499--7519}.
	\bibitem[{Chan and Elsheikh(2019)}]{RN31}
	\bibinfo{author}{Chan, S.}, \bibinfo{author}{Elsheikh, A.H.},
	\bibinfo{year}{2019}.
	\newblock \bibinfo{title}{Parametric generation of conditional geological
		realizations using generative neural networks}.
	\newblock \bibinfo{journal}{Computational Geosciences} \bibinfo{volume}{23},
	\bibinfo{pages}{925--952}.
	\bibitem[{Darvell(2021)}]{RN32}
	\bibinfo{author}{Darvell, L.E.}, \bibinfo{year}{2021}.
	\newblock \bibinfo{title}{Drone-Derived SfM Photogrammetry and Digital Rock
		Mass Mapping of Rock Slopes}.
	\newblock \bibinfo{type}{Thesis}.
	\bibitem[{Dershowitz and Einstein(1988)}]{RN52}
	\bibinfo{author}{Dershowitz, W.S.}, \bibinfo{author}{Einstein, H.H.},
	\bibinfo{year}{1988}.
	\newblock \bibinfo{title}{Characterizing rock joint geometry with joint system
		models}.
	\newblock \bibinfo{journal}{Rock Mechanics and Rock Engineering}
	\bibinfo{volume}{21}, \bibinfo{pages}{21--51}.
	\newblock \URLprefix \url{https://doi.org/10.1007/BF01019674},
	\DOIprefix\doi{10.1007/BF01019674}.
	\bibitem[{Dong et~al.(2019)Dong, Yi and Feng}]{RN44}
	\bibinfo{author}{Dong, S.}, \bibinfo{author}{Yi, X.}, \bibinfo{author}{Feng,
		W.}, \bibinfo{year}{2019}.
	\newblock \bibinfo{title}{Quantitative evaluation and classification method of
		the cataclastic texture rock mass based on the structural plane network
		simulation}.
	\newblock \bibinfo{journal}{Rock Mechanics and Rock Engineering}
	\bibinfo{volume}{52}, \bibinfo{pages}{1767--1780}.
	\newblock \URLprefix \url{https://doi.org/10.1007/s00603-018-1635-6},
	\DOIprefix\doi{10.1007/s00603-018-1635-6}.
	\bibitem[{Einstein and Baecher(1983)}]{RN64}
	\bibinfo{author}{Einstein, H.H.}, \bibinfo{author}{Baecher, G.B.},
	\bibinfo{year}{1983}.
	\newblock \bibinfo{title}{Probabilistic and statistical methods in engineering
		geology: Specific methods and examples part i: Exploration}.
	\newblock \bibinfo{journal}{Rock mechanics and Rock engineering}
	\bibinfo{volume}{16}, \bibinfo{pages}{39--72}.
	\bibitem[{Erharter(2024)}]{RN58}
	\bibinfo{author}{Erharter, G.H.}, \bibinfo{year}{2024}.
	\newblock \bibinfo{title}{Rock mass structure characterization considering
		finite and folded discontinuities: A parametric study}.
	\newblock \bibinfo{journal}{Rock Mechanics and Rock Engineering}
	\bibinfo{volume}{57}, \bibinfo{pages}{5229--5249}.
	\newblock \URLprefix \url{https://doi.org/10.1007/s00603-024-03787-9},
	\DOIprefix\doi{10.1007/s00603-024-03787-9}.
	\bibitem[{Erharter et~al.(2024)Erharter, Bar, Hansen, Jain and Marcher}]{RN41}
	\bibinfo{author}{Erharter, G.H.}, \bibinfo{author}{Bar, N.},
	\bibinfo{author}{Hansen, T.F.}, \bibinfo{author}{Jain, S.},
	\bibinfo{author}{Marcher, T.}, \bibinfo{year}{2024}.
	\newblock \bibinfo{title}{International distribution and development of rock
		mass classification: A review}.
	\newblock \bibinfo{journal}{Rock Mechanics and Rock Engineering} \URLprefix
	\url{https://doi.org/10.1007/s00603-024-04215-8},
	\DOIprefix\doi{10.1007/s00603-024-04215-8}.
	\bibitem[{Gómez et~al.(2023)Gómez, Sanchidrián, Segarra and
		Bernardini}]{RN61}
	\bibinfo{author}{Gómez, S.}, \bibinfo{author}{Sanchidrián, J.A.},
	\bibinfo{author}{Segarra, P.}, \bibinfo{author}{Bernardini, M.},
	\bibinfo{year}{2023}.
	\newblock \bibinfo{title}{A non-parametric discrete fracture network model}.
	\newblock \bibinfo{journal}{Rock Mechanics and Rock Engineering}
	\bibinfo{volume}{56}, \bibinfo{pages}{3255--3278}.
	\newblock \URLprefix \url{https://doi.org/10.1007/s00603-022-03194-y},
	\DOIprefix\doi{10.1007/s00603-022-03194-y}.
	\bibitem[{Goodfellow et~al.()Goodfellow, Pouget-Abadie, Mirza, Xu,
		Warde-Farley, Ozair, Courville and Bengio}]{RN40}
	\bibinfo{author}{Goodfellow, I.J.}, \bibinfo{author}{Pouget-Abadie, J.},
	\bibinfo{author}{Mirza, M.}, \bibinfo{author}{Xu, B.},
	\bibinfo{author}{Warde-Farley, D.}, \bibinfo{author}{Ozair, S.},
	\bibinfo{author}{Courville, A.}, \bibinfo{author}{Bengio, Y.}, .
	\newblock \bibinfo{title}{Generative adversarial nets}, in:
	\bibinfo{booktitle}{Advances in neural information processing systems}.
	\bibitem[{Gorishniy et~al.(2021)Gorishniy, Rubachev, Khrulkov and
		Babenko}]{RN71}
	\bibinfo{author}{Gorishniy, Y.}, \bibinfo{author}{Rubachev, I.},
	\bibinfo{author}{Khrulkov, V.}, \bibinfo{author}{Babenko, A.},
	\bibinfo{year}{2021}.
	\newblock \bibinfo{title}{Revisiting deep learning models for tabular data}.
	\newblock \bibinfo{journal}{Advances in neural information processing systems}
	\bibinfo{volume}{34}, \bibinfo{pages}{18932--18943}.
	\bibitem[{Gottron and Henk(2021)}]{RN62}
	\bibinfo{author}{Gottron, D.}, \bibinfo{author}{Henk, A.},
	\bibinfo{year}{2021}.
	\newblock \bibinfo{title}{Upscaling of fractured rock mass properties – an
		example comparing discrete fracture network (dfn) modeling and empirical
		relations based on engineering rock mass classifications}.
	\newblock \bibinfo{journal}{Engineering Geology} \bibinfo{volume}{294},
	\bibinfo{pages}{106382}.
	\newblock \URLprefix
	\url{https://www.sciencedirect.com/science/article/pii/S0013795221003938},
	\DOIprefix\doi{https://doi.org/10.1016/j.enggeo.2021.106382}.
	\bibitem[{Grinsztajn et~al.(2022)Grinsztajn, Oyallon and Varoquaux}]{RN25}
	\bibinfo{author}{Grinsztajn, L.}, \bibinfo{author}{Oyallon, E.},
	\bibinfo{author}{Varoquaux, G.}, \bibinfo{year}{2022}.
	\newblock \bibinfo{title}{Why do tree-based models still outperform deep
		learning on typical tabular data?}
	\newblock \bibinfo{journal}{Advances in neural information processing systems}
	\bibinfo{volume}{35}, \bibinfo{pages}{507--520}.
	\bibitem[{Han et~al.(2016)Han, Chen, Wang, Li, Zhang and Yu}]{RN22}
	\bibinfo{author}{Han, X.}, \bibinfo{author}{Chen, J.}, \bibinfo{author}{Wang,
		Q.}, \bibinfo{author}{Li, Y.}, \bibinfo{author}{Zhang, W.},
	\bibinfo{author}{Yu, T.}, \bibinfo{year}{2016}.
	\newblock \bibinfo{title}{A 3d fracture network model for the undisturbed rock
		mass at the songta dam site based on small samples}.
	\newblock \bibinfo{journal}{Rock Mechanics and Rock Engineering}
	\bibinfo{volume}{49}, \bibinfo{pages}{611--619}.
	\newblock \URLprefix \url{https://doi.org/10.1007/s00603-015-0747-5},
	\DOIprefix\doi{10.1007/s00603-015-0747-5}.
	\bibitem[{Harshvardhan et~al.(2020)Harshvardhan, Gourisaria, Pandey and
		Rautaray}]{RN68}
	\bibinfo{author}{Harshvardhan, G.}, \bibinfo{author}{Gourisaria, M.K.},
	\bibinfo{author}{Pandey, M.}, \bibinfo{author}{Rautaray, S.S.},
	\bibinfo{year}{2020}.
	\newblock \bibinfo{title}{A comprehensive survey and analysis of generative
		models in machine learning}.
	\newblock \bibinfo{journal}{Computer Science Review} \bibinfo{volume}{38},
	\bibinfo{pages}{100285}.
	\bibitem[{Ho et~al.()Ho, Jain and Abbeel}]{RN39}
	\bibinfo{author}{Ho, J.}, \bibinfo{author}{Jain, A.}, \bibinfo{author}{Abbeel,
		P.}, .
	\newblock \bibinfo{title}{Denoising diffusion probabilistic models}, in:
	\bibinfo{booktitle}{Advances in neural information processing systems}, pp.
	\bibinfo{pages}{6840--6851}.
	\bibitem[{Hollmann et~al.(2025)Hollmann, Müller, Purucker, Krishnakumar,
		Körfer, Hoo, Schirrmeister and Hutter}]{RN23}
	\bibinfo{author}{Hollmann, N.}, \bibinfo{author}{Müller, S.},
	\bibinfo{author}{Purucker, L.}, \bibinfo{author}{Krishnakumar, A.},
	\bibinfo{author}{Körfer, M.}, \bibinfo{author}{Hoo, S.B.},
	\bibinfo{author}{Schirrmeister, R.T.}, \bibinfo{author}{Hutter, F.},
	\bibinfo{year}{2025}.
	\newblock \bibinfo{title}{Accurate predictions on small data with a tabular
		foundation model}.
	\newblock \bibinfo{journal}{Nature} \bibinfo{volume}{637},
	\bibinfo{pages}{319--326}.
	\newblock \URLprefix \url{https://doi.org/10.1038/s41586-024-08328-6},
	\DOIprefix\doi{10.1038/s41586-024-08328-6}.
	\bibitem[{Huang et~al.(2017)Huang, Kulatilake, Shreedharan, Cai and
		Song}]{RN55}
	\bibinfo{author}{Huang, G.}, \bibinfo{author}{Kulatilake, P.H.S.W.},
	\bibinfo{author}{Shreedharan, S.}, \bibinfo{author}{Cai, S.},
	\bibinfo{author}{Song, H.}, \bibinfo{year}{2017}.
	\newblock \bibinfo{title}{3-d discontinuum numerical modeling of subsidence
		incorporating ore extraction and backfilling operations in an underground
		iron mine in china}.
	\newblock \bibinfo{journal}{International Journal of Mining Science and
		Technology} \bibinfo{volume}{27}, \bibinfo{pages}{191--201}.
	\newblock \URLprefix
	\url{https://www.sciencedirect.com/science/article/pii/S2095268617300617},
	\DOIprefix\doi{https://doi.org/10.1016/j.ijmst.2017.01.015}.
	\bibitem[{Illeditsch and Preh(2024)}]{RN59}
	\bibinfo{author}{Illeditsch, M.}, \bibinfo{author}{Preh, A.},
	\bibinfo{year}{2024}.
	\newblock \bibinfo{title}{Determination of meaningful block sizes for rockfall
		modelling}.
	\newblock \bibinfo{journal}{Natural Hazards} \bibinfo{volume}{120},
	\bibinfo{pages}{5685--5710}.
	\newblock \URLprefix \url{https://doi.org/10.1007/s11069-024-06432-4},
	\DOIprefix\doi{10.1007/s11069-024-06432-4}.
	\bibitem[{LeCun et~al.(2015)LeCun, Bengio and Hinton}]{RN69}
	\bibinfo{author}{LeCun, Y.}, \bibinfo{author}{Bengio, Y.},
	\bibinfo{author}{Hinton, G.}, \bibinfo{year}{2015}.
	\newblock \bibinfo{title}{Deep learning}.
	\newblock \bibinfo{journal}{nature} \bibinfo{volume}{521},
	\bibinfo{pages}{436--444}.
	\bibitem[{Li et~al.(2019)Li, Li, Li, Zhang and Zhou}]{RN45}
	\bibinfo{author}{Li, H.b.}, \bibinfo{author}{Li, X.w.}, \bibinfo{author}{Li,
		W.z.}, \bibinfo{author}{Zhang, S.l.}, \bibinfo{author}{Zhou, J.w.},
	\bibinfo{year}{2019}.
	\newblock \bibinfo{title}{Quantitative assessment for the rockfall hazard in a
		post-earthquake high rock slope using terrestrial laser scanning}.
	\newblock \bibinfo{journal}{Engineering Geology} \bibinfo{volume}{248},
	\bibinfo{pages}{1--13}.
	\newblock \URLprefix
	\url{https://www.sciencedirect.com/science/article/pii/S0013795217318835},
	\DOIprefix\doi{https://doi.org/10.1016/j.enggeo.2018.11.003}.
	\bibitem[{Li et~al.(2017)Li, Chen and Shang}]{RN43}
	\bibinfo{author}{Li, Y.}, \bibinfo{author}{Chen, J.}, \bibinfo{author}{Shang,
		Y.}, \bibinfo{year}{2017}.
	\newblock \bibinfo{title}{Connectivity of three-dimensional fracture networks:
		A case study from a dam site in southwest china}.
	\newblock \bibinfo{journal}{Rock Mechanics and Rock Engineering}
	\bibinfo{volume}{50}, \bibinfo{pages}{241--249}.
	\newblock \URLprefix \url{https://doi.org/10.1007/s00603-016-1062-5},
	\DOIprefix\doi{10.1007/s00603-016-1062-5}.
	\bibitem[{Liu et~al.(2025)Liu, Chen and Zhan}]{RN12}
	\bibinfo{author}{Liu, Y.}, \bibinfo{author}{Chen, J.}, \bibinfo{author}{Zhan,
		J.}, \bibinfo{year}{2025}.
	\newblock \bibinfo{title}{Revisiting each fracture size and spatial pattern:
		Inference from rock mass surface to interior}.
	\newblock \bibinfo{journal}{Journal of Rock Mechanics and Geotechnical
		Engineering} \bibinfo{volume}{17}, \bibinfo{pages}{1399--1417}.
	\bibitem[{Ma et~al.(2025)Ma, Zhang, Yang, Liu, Mu and Zhong}]{RN42}
	\bibinfo{author}{Ma, Q.}, \bibinfo{author}{Zhang, P.}, \bibinfo{author}{Yang,
		T.}, \bibinfo{author}{Liu, X.}, \bibinfo{author}{Mu, W.},
	\bibinfo{author}{Zhong, J.}, \bibinfo{year}{2025}.
	\newblock \bibinfo{title}{Deformation and damage mechanisms of y-intersecting
		jointed rocks under uniaxial compression}.
	\newblock \bibinfo{journal}{International Journal of Mining Science and
		Technology} \bibinfo{volume}{35}, \bibinfo{pages}{817--835}.
	\newblock \URLprefix
	\url{https://www.sciencedirect.com/science/article/pii/S2095268625000734},
	\DOIprefix\doi{https://doi.org/10.1016/j.ijmst.2025.04.005}.
	\bibitem[{Ma et~al.(2024)Ma, Mei and Xu}]{RN70}
	\bibinfo{author}{Ma, Z.}, \bibinfo{author}{Mei, G.}, \bibinfo{author}{Xu, N.},
	\bibinfo{year}{2024}.
	\newblock \bibinfo{title}{Generative deep learning for data generation in
		natural hazard analysis: motivations, advances, challenges, and
		opportunities}.
	\newblock \bibinfo{journal}{Artificial Intelligence Review}
	\bibinfo{volume}{57}, \bibinfo{pages}{160}.
	\bibitem[{Meng et~al.(2023)Meng, Mei, Qi, Xu and Peng}]{RN18}
	\bibinfo{author}{Meng, H.}, \bibinfo{author}{Mei, G.}, \bibinfo{author}{Qi,
		X.}, \bibinfo{author}{Xu, N.}, \bibinfo{author}{Peng, J.},
	\bibinfo{year}{2023}.
	\newblock \bibinfo{title}{Generating stochastic structural planes by
		considering parameter correlations using deep generative adversarial
		networks}.
	\newblock \bibinfo{journal}{Rock Mechanics and Rock Engineering}
	\bibinfo{volume}{56}, \bibinfo{pages}{9215--9230}.
	\newblock \URLprefix \url{https://doi.org/10.1007/s00603-023-03553-3},
	\DOIprefix\doi{10.1007/s00603-023-03553-3}.
	\bibitem[{Meng et~al.(2024)Meng, Mei, Qi, Xu and Peng}]{RN19}
	\bibinfo{author}{Meng, H.}, \bibinfo{author}{Mei, G.}, \bibinfo{author}{Qi,
		X.}, \bibinfo{author}{Xu, N.}, \bibinfo{author}{Peng, J.},
	\bibinfo{year}{2024}.
	\newblock \bibinfo{title}{Deep generative model-based generation method of
		stochastic structural planes of rock masses in tunnels}.
	\newblock \bibinfo{journal}{Geological Journal} \bibinfo{volume}{59},
	\bibinfo{pages}{2566--2583}.
	\newblock \URLprefix \url{https://doi.org/10.1002/gj.5000},
	\DOIprefix\doi{https://doi.org/10.1002/gj.5000}.
	\bibitem[{Pan et~al.(2019)Pan, Li, Xu, Zhang, Lin and Li}]{RN48}
	\bibinfo{author}{Pan, D.}, \bibinfo{author}{Li, S.}, \bibinfo{author}{Xu, Z.},
	\bibinfo{author}{Zhang, Y.}, \bibinfo{author}{Lin, P.}, \bibinfo{author}{Li,
		H.}, \bibinfo{year}{2019}.
	\newblock \bibinfo{title}{A deterministic-stochastic identification and
		modelling method of discrete fracture networks using laser scanning:
		Development and case study}.
	\newblock \bibinfo{journal}{Engineering Geology} \bibinfo{volume}{262},
	\bibinfo{pages}{105310}.
	\newblock \URLprefix
	\url{https://www.sciencedirect.com/science/article/pii/S0013795219305897},
	\DOIprefix\doi{https://doi.org/10.1016/j.enggeo.2019.105310}.
	\bibitem[{Pan and Yang(2009)}]{RN27}
	\bibinfo{author}{Pan, S.J.}, \bibinfo{author}{Yang, Q.}, \bibinfo{year}{2009}.
	\newblock \bibinfo{title}{A survey on transfer learning}.
	\newblock \bibinfo{journal}{IEEE Transactions on knowledge and data
		engineering} \bibinfo{volume}{22}, \bibinfo{pages}{1345--1359}.
	\bibitem[{Priest(1993)}]{RN53}
	\bibinfo{author}{Priest, S.D.}, \bibinfo{year}{1993}.
	\newblock \bibinfo{title}{Discontinuity analysis for rock engineering}.
	\newblock \bibinfo{publisher}{Springer Science \& Business Media}.
	\bibitem[{Shwartz-Ziv and Armon(2022)}]{RN26}
	\bibinfo{author}{Shwartz-Ziv, R.}, \bibinfo{author}{Armon, A.},
	\bibinfo{year}{2022}.
	\newblock \bibinfo{title}{Tabular data: Deep learning is not all you need}.
	\newblock \bibinfo{journal}{Information Fusion} \bibinfo{volume}{81},
	\bibinfo{pages}{84--90}.
	\bibitem[{Sohl-Dickstein et~al.()Sohl-Dickstein, Weiss, Maheswaranathan and
		Ganguli}]{RN38}
	\bibinfo{author}{Sohl-Dickstein, J.}, \bibinfo{author}{Weiss, E.},
	\bibinfo{author}{Maheswaranathan, N.}, \bibinfo{author}{Ganguli, S.}, .
	\newblock \bibinfo{title}{Deep unsupervised learning using nonequilibrium
		thermodynamics}, in: \bibinfo{booktitle}{International conference on machine
		learning}, \bibinfo{publisher}{pmlr}. pp. \bibinfo{pages}{2256--2265}.
	\bibitem[{Sun et~al.(2025)Sun, Zheng, Xie, Liu, Chu, Qiu, Xu, Ding, Li and
		Geng}]{RN73}
	\bibinfo{author}{Sun, J.}, \bibinfo{author}{Zheng, C.}, \bibinfo{author}{Xie,
		E.}, \bibinfo{author}{Liu, Z.}, \bibinfo{author}{Chu, R.},
	\bibinfo{author}{Qiu, J.}, \bibinfo{author}{Xu, J.}, \bibinfo{author}{Ding,
		M.}, \bibinfo{author}{Li, H.}, \bibinfo{author}{Geng, M.},
	\bibinfo{year}{2025}.
	\newblock \bibinfo{title}{A survey of reasoning with foundation models:
		Concepts, methodologies, and outlook}.
	\newblock \bibinfo{journal}{ACM Computing Surveys} \bibinfo{volume}{57},
	\bibinfo{pages}{1--43}.
	\bibitem[{Wang et~al.(2018)Wang, Li, Qiao, Liu and Yang}]{RN66}
	\bibinfo{author}{Wang, Z.}, \bibinfo{author}{Li, W.}, \bibinfo{author}{Qiao,
		L.}, \bibinfo{author}{Liu, J.}, \bibinfo{author}{Yang, J.},
	\bibinfo{year}{2018}.
	\newblock \bibinfo{title}{Hydraulic properties of fractured rock mass with
		correlated fracture length and aperture in both radial and unidirectional
		flow configurations}.
	\newblock \bibinfo{journal}{Computers and Geotechnics} \bibinfo{volume}{104},
	\bibinfo{pages}{167--184}.
	\bibitem[{Yi et~al.(2025)Yi, Feng, Zhou, Wu and Zhao}]{RN20}
	\bibinfo{author}{Yi, X.}, \bibinfo{author}{Feng, W.}, \bibinfo{author}{Zhou,
		Y.}, \bibinfo{author}{Wu, W.}, \bibinfo{author}{Zhao, J.},
	\bibinfo{year}{2025}.
	\newblock \bibinfo{title}{Uncertainty quantification of rock mass discontinuity
		information and mechanical parameter derived from photogrammetry and monte
		carlo simulation: A case study}.
	\newblock \bibinfo{journal}{Results in Earth Sciences} \bibinfo{volume}{3},
	\bibinfo{pages}{100060}.
	\newblock \URLprefix
	\url{https://www.sciencedirect.com/science/article/pii/S2211714825000020},
	\DOIprefix\doi{https://doi.org/10.1016/j.rines.2025.100060}.
	
	
	\bibitem[{Zanbak(1977)}]{RN67}
	\bibinfo{author}{Zanbak, C.},
	\newblock \bibinfo{year}{1977}.
	\newblock \bibinfo{title}{Statistical interpretation of discontinuity contour diagrams},
	\bibinfo{publisher}{Elsevier}. 
	

	\bibitem[{Zhang et~al.(2023)Zhang, Wu, Zhang and Huang}]{RN46}
	\bibinfo{author}{Zhang, H.}, \bibinfo{author}{Wu, S.}, \bibinfo{author}{Zhang,
		Z.}, \bibinfo{author}{Huang, S.}, \bibinfo{year}{2023}.
	\newblock \bibinfo{title}{Reliability analysis of rock slopes considering the
		uncertainty of joint spatial distributions}.
	\newblock \bibinfo{journal}{Computers and Geotechnics} \bibinfo{volume}{161},
	\bibinfo{pages}{105566}.
	\newblock \URLprefix
	\url{https://www.sciencedirect.com/science/article/pii/S0266352X23003233},
	\DOIprefix\doi{https://doi.org/10.1016/j.compgeo.2023.105566}.
	
	\bibitem[{Priest and Hudson(1981)}]{RN1981}
	\bibinfo{author}{Priest, S.~D.}, \bibinfo{author}{Hudson, J.~A.}, \bibinfo{year}{1981}.
	\newblock \bibinfo{title}{Estimation of discontinuity spacing and trace length using scanline surveys}.
	\newblock \bibinfo{journal}{International Journal of Rock Mechanics and Mining Sciences \& Geomechanics Abstracts} \bibinfo{volume}{18},
	\bibinfo{number}{3}, \bibinfo{pages}{183--197}.
	
	\bibitem[{Zhang et~al.(2019)Zhang, Tilke, Dupont, Zhu, Liang and Bailey}]{RN30}
	\bibinfo{author}{Zhang, T.F.}, \bibinfo{author}{Tilke, P.},
	\bibinfo{author}{Dupont, E.}, \bibinfo{author}{Zhu, L.C.},
	\bibinfo{author}{Liang, L.}, \bibinfo{author}{Bailey, W.},
	\bibinfo{year}{2019}.
	\newblock \bibinfo{title}{Generating geologically realistic 3d reservoir facies
		models using deep learning of sedimentary architecture with generative
		adversarial networks}.
	\newblock \bibinfo{journal}{Petroleum Science} \bibinfo{volume}{16},
	\bibinfo{pages}{541--549}.
	\bibitem[{Zhang and Lei(2013)}]{RN65}
	\bibinfo{author}{Zhang, Z.}, \bibinfo{author}{Lei, Q.}, \bibinfo{year}{2013}.
	\newblock \bibinfo{title}{Object-oriented modeling for three-dimensional
		multi-block systems}.
	\newblock \bibinfo{journal}{Computers Geotechnics} \bibinfo{volume}{48},
	\bibinfo{pages}{208--227}.
	\bibitem[{Zheng and Zhang(2022)}]{RN17}
	\bibinfo{author}{Zheng, Q.}, \bibinfo{author}{Zhang, D.}, \bibinfo{year}{2022}.
	\newblock \bibinfo{title}{Digital rock reconstruction with user-defined
		properties using conditional generative adversarial networks}.
	\newblock \bibinfo{journal}{Transport in Porous Media} \bibinfo{volume}{144},
	\bibinfo{pages}{255--281}.
	\newblock \URLprefix \url{https://doi.org/10.1007/s11242-021-01728-6},
	\DOIprefix\doi{10.1007/s11242-021-01728-6}.
	
\end{thebibliography}


\end{document}